\documentclass{article}

\usepackage{arxiv}
\usepackage{svg}
\usepackage{multirow}
\usepackage[utf8]{inputenc} 
\usepackage[T1]{fontenc}    
\usepackage{hyperref}       
\usepackage{url}            
\usepackage{booktabs}       
\usepackage{amsfonts}       
\usepackage{nicefrac}       
\usepackage{microtype}      
\usepackage{lipsum}
\usepackage{graphicx}
\usepackage{amsmath,graphicx}
\usepackage{cite}
\usepackage{balance}
\usepackage{enumitem}
\usepackage{flushend}

\graphicspath{ {./images/} }

\title{Optimal Transport-based Graph Matching for 3D retinal OCT image registration}

\author{
 Xin Tian \\
  Department of Electrical and Electronic Engineering\\
  University of Bristol\\
  Bristol BS8 1UB, UK \\
  \texttt{xin.tian@bristol.ac.uk} \\
   \And
  Nantheera Anantrasirichai\\
  School of Cellular and Molecular Medicine\\
  University of Bristol\\
  Bristol BS8 1TD, UK \\
  \texttt{N.Anantrasirichai@bristol.ac.uk} \\
    \And
  Lindsay Nicholson\\
  Department of Electrical and Electronic Engineering\\
  University of Bristol\\
  Bristol BS8 1UB, UK \\
  \texttt{L.Nicholson@bristol.ac.uk} \\
  \And
 Alin Achim \\
  Department of Electrical and Electronic Engineering\\
  University of Bristol\\
  Bristol BS8 1UB, UK \\
  \texttt{alin.achim@bristol.ac.uk} \\
}

\begin{document}

\maketitle

\begin{abstract}

\noindent Registration of longitudinal optical coherence tomography (OCT) images assists disease monitoring and is essential in image fusion applications. Mouse retinal OCT images are often collected for longitudinal study of eye disease models such as uveitis, but their quality is often poor compared with human imaging. This paper presents a novel but efficient framework involving an optimal transport based graph matching (OT-GM) method for 3D mouse OCT image registration. We first perform registration of fundus-like images obtained by projecting all b-scans of a volume on a plane orthogonal to them, hereafter referred to as the x-y plane. We introduce Adaptive Weighted Vessel Graph Descriptors (AWVGD) and 3D Cube Descriptors (CD) to identify the correspondence between nodes of graphs extracted from segmented vessels within the OCT projection images. The AWVGD comprises scaling, translation and rotation, which are computationally efficient, whereas CD exploits 3D spatial and frequency domain information. The OT-GM method subsequently performs the correct alignment in the x-y plane. Finally, registration along the direction orthogonal to the x-y plane (the z-direction) is guided by the segmentation of two important anatomical features peculiar to mouse b-scans, the Internal Limiting Membrane (ILM) and the hyaloid remnant (HR). Both subjective and objective evaluation results demonstrate that our framework outperforms other well-established methods on mouse OCT images within a reasonable execution time.
\end{abstract}

\section{Introduction}
\label{sec:intro}

 Retinal image registration aims to spatially align two or more retinal images captured at different times and from different viewpoints for clinical review of disease progression. Nowadays, non-invasive Optical Coherence Tomography (OCT), which enables 3-D retinal  imaging, is widely used for in-vivo assessment of eye disease\cite{retinareview2010, 6556778}. Data is generated by acquiring a series of A-scans that are combined to a 2D cross-sectional slice called a B scan. The summation of B-scans creates a fundus like projection image. 

An efficient longitudinal mouse OCT registration method has the potential to impact both early studies of disease mechanisms and treatment. It also helps generating strategies for analysis that are translatable to human imaging. However, imaging the mouse retina is more challenging and less well studied than is the case in humans. High-quality mouse OCT is usually expensive and hard to acquire. Low quality datasets suffer from major vascular dissimilarities, distortions, noise, intensity inconsistency and blurring effects, making the registration highly challenging. Deep learning methods are also infeasible because ground truth is unavailable. 
 
The registration of longitudinal OCT volumes requires correction of displacement in the entire x-y-z volume. Some methods register the B-scans directly (x-z planes, see Fig. \ref{fig:flow}), such as the spatially region-weighted correlation ratio (SRWCR) based method \cite{SRWCR}. However, the registration result could be wrong without alignment in the x and y directions.
A more commonly used strategy for longitudinal OCT volume registration is to apply 2D projection registration for x-y direction displacements correction, followed by registering the B-scans in the z-direction. The OCT projection registration is usually performed by adapting methods from fundus registration. In \cite{GDB}, edge and corner key points within a growing bootstrap region are used to improve an initial transformation obtained from SIFT\cite{SIFT}. A non-rigid fundus registration method is generally applied, e.g. Gaussian Field Estimator with Manifold Regularization (GFEMR) \cite{GFEMR}. The Harris-PIIFD \cite{PIIFD} integrates the Harris-Stephen corner detector gradient-based filters and image histograms to produce a set of Partial Intensity Invariant Feature Descriptors (PIIFD).  These methods all exhibit robustness problems. Meanwhile, the Optimal Transport (OT) theory emerged as a robust and flexible tool to address complex problems in many domains \cite{Papadakis2015, Haker2004, Zhu2007}. Motta et al. \cite{VOTUS} proposed a fully automated OT-driven graph matching framework for human fundus registration.

Pan et al. \cite{seg3D} and Chen et al.\cite{OCTRexpert} proposed layer segmentation guided 3D OCT registration and achieve good performance for human datasets. However, segmentation methods adapted from human OCT are often limited in the mouse model because of segmentation errors. The existence of the hyaloid remnant (HR) and differences in the OCT reflection profile between the human and mouse retina make existing layer segmentation algorithms for B-scan less efficient \cite{mouseseg}. Although well-established for human retina image registration, the existing algorithms cannot register well poor quality mouse OCT image pairs. The major vascular dissimilarities, noise, severe intensity inconsistency make the key points detection and descriptors extraction difficult. The different retina structure also makes layer segmentation guided methods less adequate for mouse images.

In this paper, we present a novel framework for 3D mouse OCT image registration. This involves an optimal transport based graph matching (OT-GM) method with robust descriptors to address the difficulties mentioned above. The remainder of this paper is organised as mentioned in the following. Section 2 devoted to the proposed OT-GM based x-y registration with our novel Adaptive Weighted Vessel Graph Descriptors (AWVGD) and 3D Cube Descriptors (CD), and anatomical features segmentation guided z-direction registration. The AWVGD comprises scaling, translation and rotation, which are computationally efficient. In Section 3, the experiment results are subjectively and objectively evaluated and analysed with the comparison with other methods. Finally, conclusions and future work are presented in Section 4.

\section{PROPOSED REGISTRATION FRAMEWORK}
\label{sec:method}
\begin{figure*}[ht]
\includegraphics[width=16.5cm]{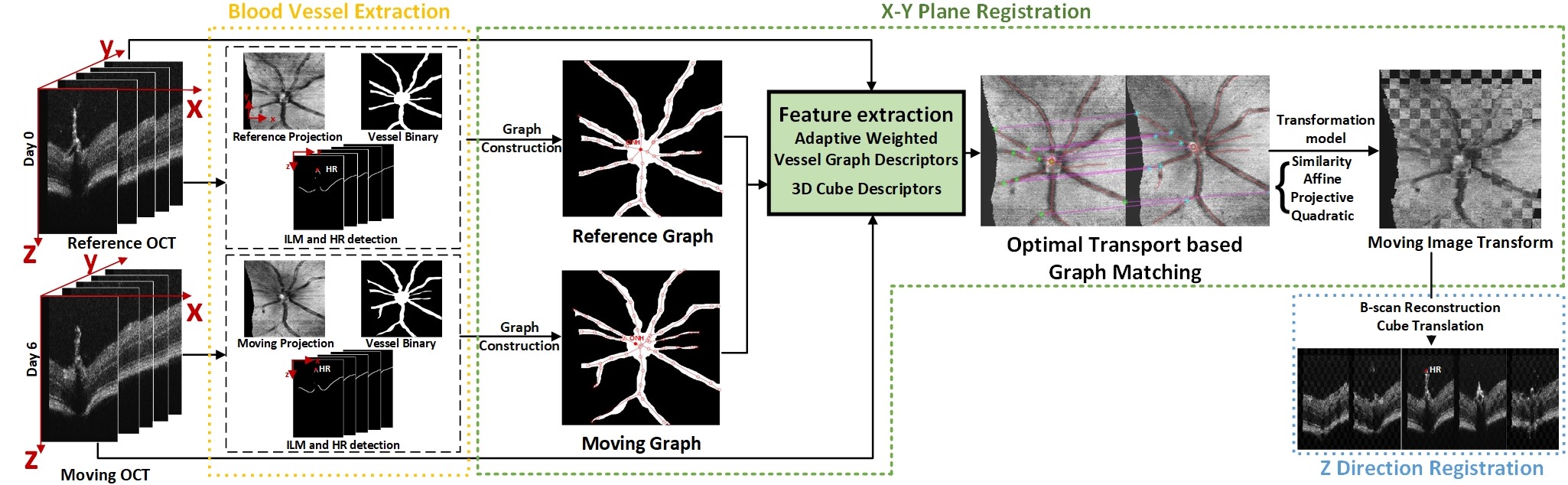}
\caption{The flowchart of the proposed 3D mouse OCT registration.}
\label{fig:flow}
\end{figure*}

The proposed 3D OCT registration framework involves: i) blood vessel extraction, ii) x-y plane registration, and iii) z-direction registration. The overall flowchart is shown in Fig. \ref{fig:flow}, with the B-scans defined at the x-z plane and the projection image at the x-y plane. 

\subsection{Blood vessel extraction}
\label{sec: preprocessing}
Firstly, OCT volumes captured on different days are denoised using \cite{Pui}, and the projection image is generated by summing up in the z-direction. To extract blood vessels, we employ the wavelet-based method \cite{ICASSP}, designed especially for mouse OCT projections segmentation. Based on the analysis in \cite{7351548} where the finest wavelet decomposition levels are the most sensitive to noise and blur. We used decomposition levels of 2, 3, 5, and the multiplication product produces the best average results for extracting vessel binaries of mouse projections.

The Internal Limiting Membrane (ILM) boundary is the first and most prominent layer boundary with the highest contrast in mouse OCT images. The ILM segmentation for each B-scan is completed by Otsu threshold and Canny edge detection. Then the profile of blood vessels, comprising optic nerve head (ONH) and hyaloid remnant (HR) is extracted. The HR is a branch of the ophthalmic artery that extends from the ONH through the vitreous humour to the lens. The ONH is an important anatomical feature in the projection image that represents where blood and nerve fibres converge into the retina. 
The projection of the apex of HR (the highest ILM points in B-scan volume, see Fig. \ref{fig:flow}) denotes the centroid of ONH in the projection image. The segmentation of ILM, HR apex and ONH centroid are key features for further steps. 

\begin{figure*}[t]
\begin{minipage}[b]{.5\linewidth}
  \centering
  \centerline{\includegraphics[height=5.5cm, width=3.5cm]{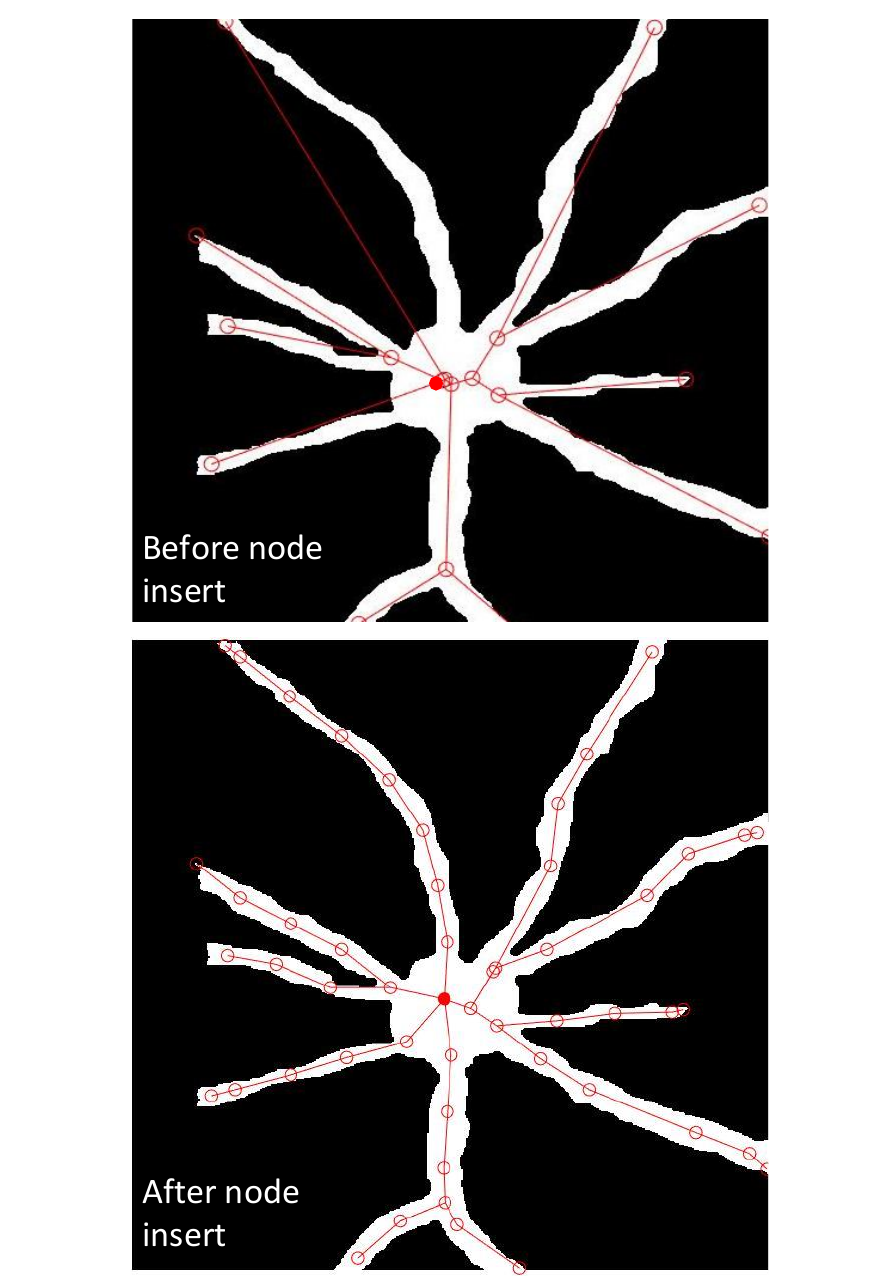}}
  \centerline{(a)}
\end{minipage}
\hfill
\begin{minipage}[b]{.6\linewidth}
  \centering
  \centerline{\includegraphics[height=5.2cm, width=5.5cm]{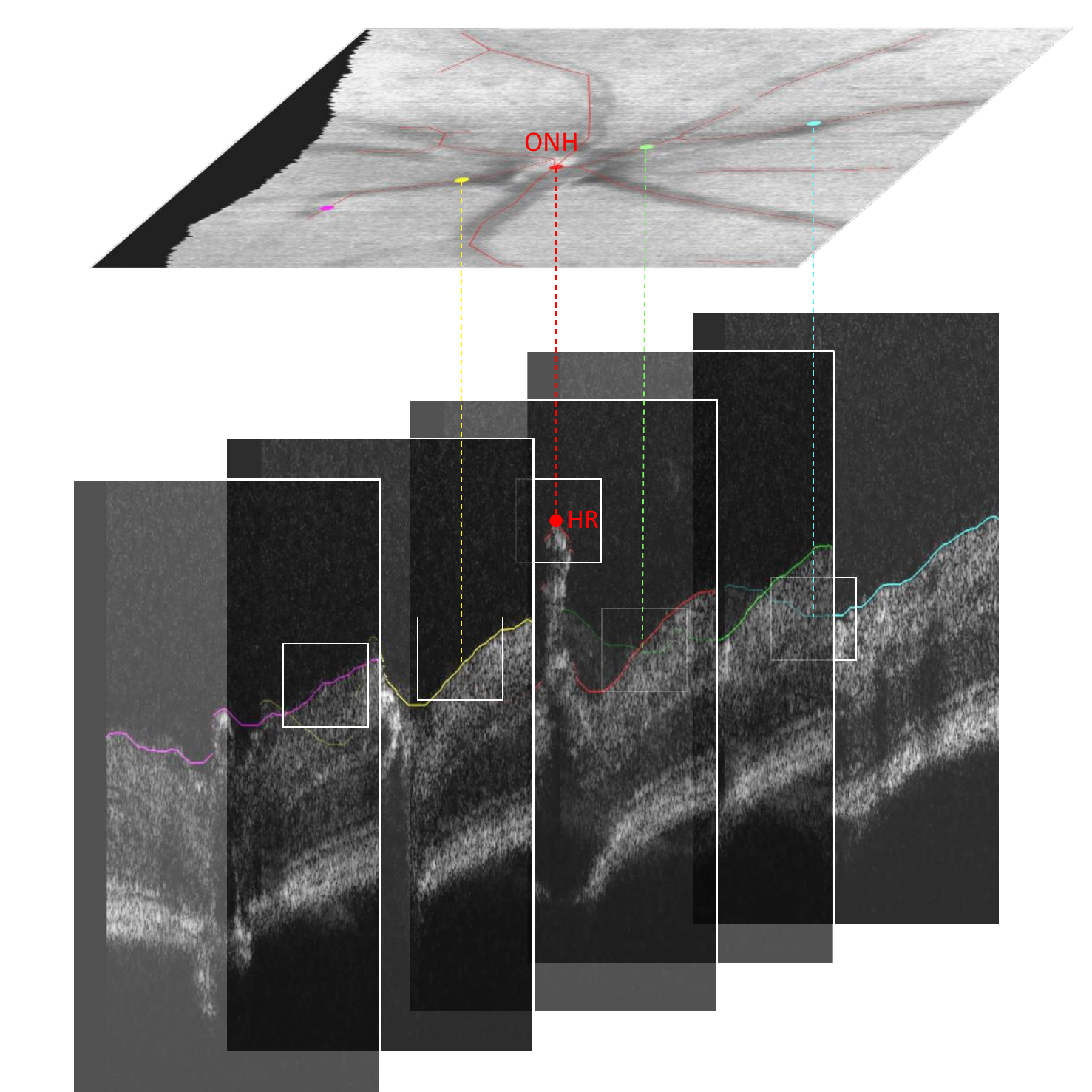}}
  \centerline{(b)}
\end{minipage}
\caption{(a) Graph construction results before (top) and after (bottom) nodes insert and merge. (b) The 3D Volume feature extraction based on the graph nodes and ILM positions.}
\label{fig:ILM}
\end{figure*}

\subsection{X-Y Plane Registration for Projection Images}
\label{sec:xy}
Our x-y registration process follows a discrete optimal transport based graph matching model (OT-GM) \cite{VOTUS}. Here, the cost function combines various similarity measures of new 3D cube descriptors (CD), more suitable for mouse OCT. We propose an Adaptive Weighted Vessel Graph Descriptors (AWVGD), which includes scaling, translation and rotation to achieve better performance while preserving computational efficiency. 
After matching, the moving image transformation is selected from similarity, affine, projective, and quadratic transformation models.

\subsubsection{Graph Construction}
Given the reference projection $R$ and moving projection $M$, let $G_R(V_R,E_R)$ and $G_M(V_M,E_M)$ be the blood vessel graphs extracted from vessel binaries.
The initial graph extraction is implemented by first extracting the branch and endpoints from the vessel's skeleton. The vessel skeleton is extracted through morphological operations, where $V_R$ and $V_M$ represent nodes of $R$ and $M$. $E_R$ and $E_M$ are edges weighted by the distance between the nodes. 
As the mouse OCT projection does not include small vessels branches, there are not enough branches and endpoints that can be extracted as key points. Thus we insert new nodes on existing vessel skeletons with the spatial distance $d$, which is selected as a trade-off between accuracy and computational speed as shown in Fig. \ref{fig:ILM} (a). $d$ is empirically set to 46 pixels. If the new node is within 10 pixels from the existing node, they are merged to avoid redundancy. The nodes are consequently well distributed along all vessels. 
\subsubsection{Optimal Transport based Graph Matching}
 Defining $X$ as a matrix of size $N_R \times N_M$, with $N_R=|V_R|$ and $N_M=|V_M|$, that matches the set of nodes $V_R$ to $V_M$, the graphs produced are matched by solving the optimal transportation plan $X$ to satisfy Equation (\ref{eg:argminX}) \cite{VOTUS}.
 
     \begin{equation*}
      \mathop {\mathrm {arg\,min}} _ {\mathbf {X}}\, \displaystyle \sum _{i = 1}^{N_{R}} \sum _{j = 1}^{N_{M}} C_{ij} \, X_{ij},
       \tag{1}
       \label{eg:argminX}
    \end{equation*}
    
    \noindent where $C_{ij}$ represents the coefficient of a cost matrix $\mathbf C$ , which accounts for the transportation cost to move a portion of mass from $\mathbf {v}_{R}^{i}$ to $\mathbf {v}_{M}^{j}$. Each coefficient $X_{ij}$ of $\mathbf X$ provides the transfer route for conveying a certain quantity of mass from the node $\mathbf {v}_{R}^{i} \in V_{R}$ to the other one $\mathbf {v}_{M}^{j} \in V_{M}$. The total mass of the system is constrained to be unchanged by $\sum _{j} X_{ij} = \mu (\mathbf {v}_{I}^{i})$ and $\sum _{i} X_{ij} = \nu (\mathbf {v}_{R}^{j})$ with $\mu (\mathbf {v}_{I}^{i}) = \nu (\mathbf {v}_{R}^{j}) = 1$. As the masses are set to unitary values, all entries of the optimal solution matrix $X$ will be binaries to allow the Mixed Integer Linear Programming (MILP) solver  \cite{MILP} to tackle the optimization problem. 
    
    To tackle mass difference and outliers between the graphs, two ghost nodes $\mathbf g_R$ and $\mathbf  g_M$ are added with mass $\alpha$ and $ N_I-N_R+ \alpha $, respectively. Quantity $N_I-N_R$ regulates the difference of masses between $G_I$ and $G_R$ , while $\alpha = \omega \min {(N_I,N_R)}, \omega \in [0,1]$ is a fixed threshold which estimates the number of outliers. The value $\omega$ depends on the number of outliers in the graphs. 
    Once the optimization is performed, all nodes matched with ghost nodes are discarded, and the correspondences between the matched nodes are used for image transformation.
    
\subsubsection{Feature Descriptors from Vessel Graph and 3D OCT Volume}
 In this paper, we define the cost matrix $\mathbf C$ in Equation (1) as: 
    \begin{equation*} 
    \mathbf C = \mathbf {C}_{G}+ \mathbf {C}_{V}, \\
    \tag{2}
    \end{equation*}

\noindent where $\mathbf {C}_{G}$ denotes the cost matrix built according to vessel graph descriptors and $\mathbf {C}_{V}$ with cube descriptors
    \begin{equation*} 
    \begin{cases}
    \mathbf {C}_{G} = \mathbf{D_{G}} + \mathbf{S_{G}} \\
    \mathbf {C}_{V} = \mathbf{D_{V}} + \mathbf{S_{V}}
    \end{cases}
    \tag{3}
    \end{equation*}
 \noindent where $\mathbf{D}$ is the matrices of node-to-node dissimilarities. The matrices $\mathbf{S}$ is neighbour-to-neighbour dissimilarities computed by exploiting the adjacency elements given by the edges. The coefficient ${D}_ij$ from matrix $\mathbf{D}$ accounts for the cost to transport a unit of mass from the a node $\mathbf{v}_{R}^{i}$ to the other node $\mathbf{v}_{M}^{j}$.The elements of  $\mathbf{D}$ and  $\mathbf{S}$ are normalized in the range $[0,1]$.

\noindent \textbf{$\mathbf{D_{G}}$: Adaptive Weighted Vessel Graph Descriptors (AWVGD) } is introduced to compute graph dissimilarities, defined as:
\begin{equation} 
D_{G}(\!\mathbf {v}\!_{R}\!^{i},\!\mathbf {v}\!_{M}^{j}\!)\! =\!\sum_{m=1}^{Y}\!{ {w}_{G_s}^{m} {\lVert \mathrm {desc}_{G}^{m}(\!\mathbf {v}_{R}^{i}\!)\!-\!\mathrm {desc}_{G}^{m}(\!\mathbf {v}_{M}^{j}\!) \rVert}^{2}},
\tag{4}
\end{equation}
    
\noindent where $\mathrm {desc_G^{m}}(\cdot)$ denotes the $m^{th}$ graph descriptors, and $Y$ is the number of graph descriptors. ${\lVert \cdot \rVert}^{2} $ is the Euclidean distance for graph descriptors, $ {w}_{G_s}^{m}$ is the weighting coefficients for each graph features that are adaptively selected by analyzing the difference between $G_R(V_R,E_R)$ and $G_M(V_M,E_M)$. 

In this paper, five graph descriptors ($Y=5)$ are applied to distinguish each node from others: i) \textit{node degree} ($\mathrm {desc_G^{1}}$) to count how many neighbours each node has, ii) \textit{closeness centrality} ($\mathrm {desc_G^{2}}$) to measure the average shortest path length between a node and all other nodes, iii) \textit{betweenness centrality} ($\mathrm {desc_G^{3}}$) to record how often a node lies on the shortest path between two other nodes, iv) \textit{eigenvector centrality} ($\mathrm {desc_G^{4}}$) to account for how important a node's neighbours are, and v) \textit{edge weights summation} ($\mathrm {desc_G^{5}}$) to encode edge importance.
The descriptors are scaling, translation and rotation invariant \cite{centralities}. The local shapes and topological relations of vessels are embedded into these graphs as the graphs are extracted from vessel skeletons. 

We introduce weighting coefficients ${w}_{G_s}$ to maximize the impact of each descriptor under different graph patterns adaptively. As vessel graphs to be matched are often not identical, when some important nodes (e.g. ones with higher degree) are missing, the value of the other descriptors will be affected. Accordingly, this paper classifies the graph patterns into $s=6$ categories. For each category, we assign a different weight ${w}_{G_s}$ empirically. The six graph pattern categories are i) The main difference between $G_R(V_R,E_R)$ and $G_R(V_R,E_R)$ is the number of low degree nodes (e.g. endpoints). ii) The two graphs have a limited number of high degree nodes difference. iii) All missing nodes are high degree nodes so that the number of connected components and edges differs from two graphs. iv) The most missing nodes in one graph are uncorrelated with nodes in the other graph. v) The number of nodes, edges, and uncorrelated nodes are significantly different, but the connected component number remains close. vi) Other patterns use equal weights. Each weighting coefficients, $\sum_{m=1}^{5}  {w}_{G_s}^{m} = 1 $. The weights are acquired from synthetic graphs. The proposed AWVGD can be used in other graph matching tasks.

 
\noindent \textbf{$\mathbf{D_{V}}$: Cube Descriptor (CD)} 
The dissimilarities with cube descriptors is computed by 
    \begin{equation*} 
    D_{V}(\!\mathbf {v}_{R}^{i},\!\mathbf {v}_{M}^{j}\!)\! =\!    \sum_{k=1}^{K}  \!\mathrm {diss}_{k} \!\left (\!{\mathrm {desc}_{V}^{k}(\!\mathbf {v}_{R}^{i}\!)\!,\! \mathrm {desc}_{V}^{k}(\!\mathbf {v}_{M}^{j}\!) }\right)
    \tag{5}
    \end{equation*}
\noindent where $\mathrm {desc_V^{k}}(\cdot)$ denotes the $k^{th}$ cube descriptor, $\mathrm {diss}_{k} (\cdot,\cdot)$) represents the dissimilarity measure for the corresponding descriptor, and $K$ is the number of cube descriptors. All cube descriptors are collected at the cubes around the voxel in the 3D volume. The voxels $\mathbf o$ are located by finding the corresponding points of nodes on ILMs as shown in Fig. \ref{fig:ILM} (b). 

In this paper, three graph descriptors ($K=3)$ are applied. The $\mathrm {desc_V^{1}}$ is the intensities of a cube centred at each evaluated voxel. For this dataset, the size of the cube is set to $19\times19\times19$, empirically. The $\mathrm {desc_V^{2}}$ is to exploit the texture information of B-scans. This descriptor takes the same cube in $\mathrm {desc_V^{1}}$ and is implemented with Gabor filters \cite{gabor}. The filter bank ranges from $0^{\circ}$ to $90^{\circ}$ with a wavelength of 4 and 8 pixels/cycle, so it returns four filtered cubes. The $\mathrm {desc_V^{3}}$ is a three-dimensional vector defined by $\mathbf {u_o=o-h}$, where $\mathbf h$ denotes the apex of HR. This vector encodes the 3D spatial cues of vessels. 
Both $\mathrm {diss}_{1}$ and $\mathrm {diss}_{2}$ in Equation (5) are the cosine distance. The $\mathrm {diss}_{3}$ uses the negative cross-correlation.

\subsubsection{Transformation Model Selection}
The transformation $T$ for the projection images are selected from both deformable and non-deformable mappings, which are i) similarity, ii) affine, iii) projective, and iv) quadratic models. This selection is guided by computing the Gain Coefficient (GC) \cite{GC} between vessel binaries after transformation as in Equation (7). It measures the amount of pixels aligned after the mapping by the geometric transformation. The more pixels are overlapped after transformation, the higher the GC. The best geometric transformation among all four models is the one with the maximum GC.
\begin{equation*} GC(B_{R},B_{M},T) = \frac {| B_{M} \cap T(B_{R}) |}{| B_{M} \cap B_{R} |} \,.\tag{7}
\end{equation*}

The GC is also used as an objective evaluation index for measuring the performance since the ground truth for mouse OCT registration is difficult to define.  

\subsection{Z Direction Registration}
As each pixel on the projection image corresponds to one A-scan of an OCT volume, transformation for projection images can be applied to individual A-scans of the subject OCT volume to reconstruct the B-scans. After B-scans are reconstructed, a coarse translation in the z-direction is calculated according to the difference of HR position between two OCT volumes. The translation is then iteratively refined for each B-scan according to the average ILM position of each slice.

\section{EXPERIMENTAL RESULT AND ANALYSIS}
\label{sec:experiment}
\noindent \textbf{Dataset:}
Our longitudinal dataset has 43 pairs of retinal OCT volumes (1024×512×512 pixels) collected from 32 mice with uveitis. The OCT images were obtained using the Micron IV fundus camera and an OCT scan head equipped with a mouse objective lens (Phoenix Technologies, California) at different time points (days 0, 2, 7, 14). Volume scans were taken around the optic disc at the centre. The dataset suffers from major vascular dissimilarities, distortions, noise, and blurring effects that makes the registration challenging.

    \noindent \textbf{Evaluation method:}
    Because of the low quality of the dataset, precise vessel and layer segmentation and control points are not available for calculating commonly used registration evaluation metrics (e.g. Dice, MSE, MRE). Thus, the registration results are generally subjectively evaluated by a bio-medicine expert visually to define if registration can be considered valid or not based on experience. For objective assessment where possible, the GC is used as an objective evaluation metric to evaluate valid registration results as shown in Table \ref{tab:regresult}. The higher the GC, the better the registration. The GC of invalid registration results are set to 0. The average execution time is also evaluated.
   
    \begin{figure}[t]
    \centering
\includegraphics[width=8.5cm]{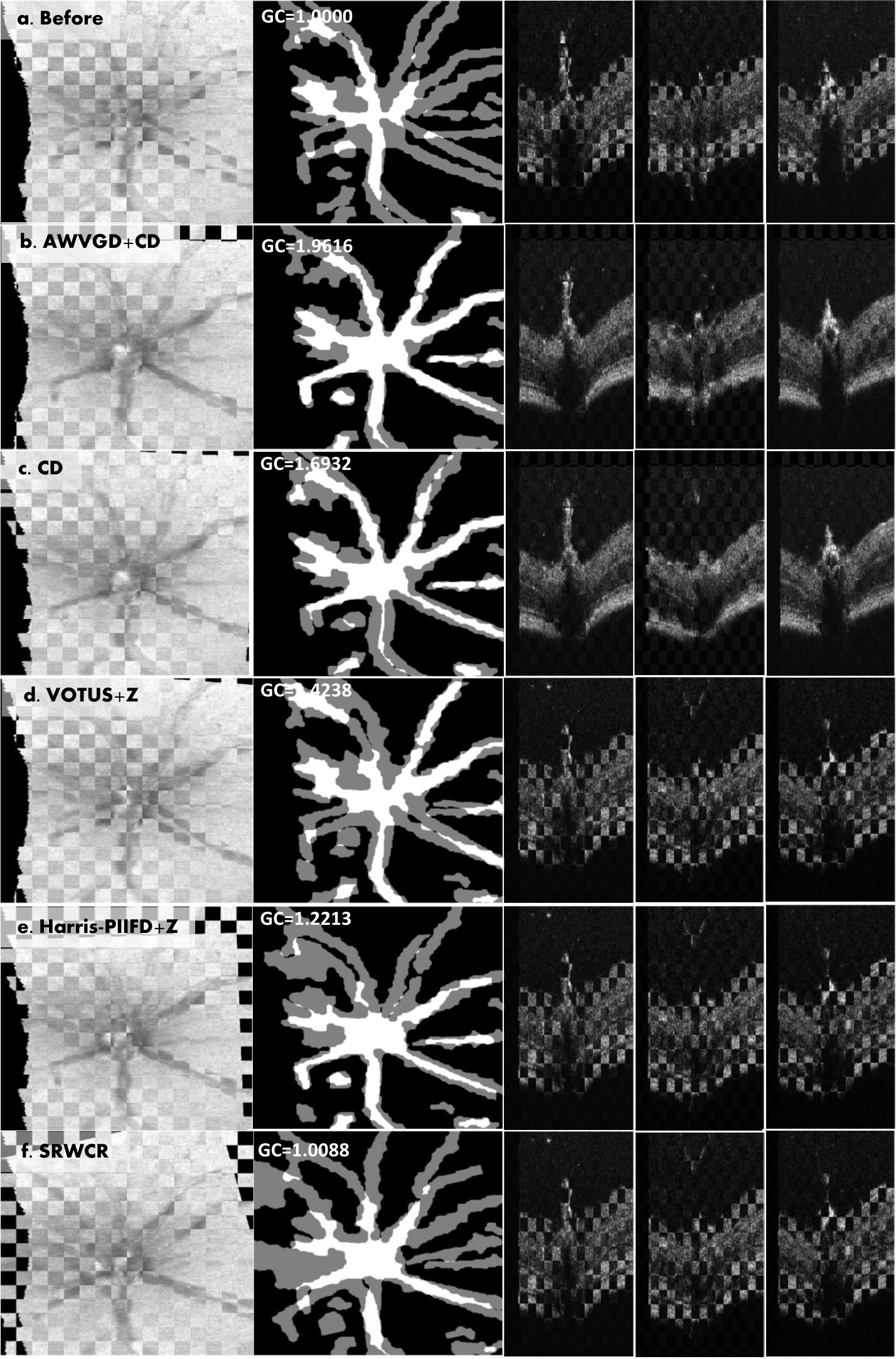}
\caption{Comparison results of different registration methods with the checkerboard of the reference and moving image, and its vessel segmentation, B-scans at ONH and two scans 60 slices away from left and right. (a) Before registration (b) our AWVDG+CD (c) CD (d) VOTUS+Z (e) Harris-PIIFD+Z (d) SRWCR.}
\label{fig:results}
\end{figure}
     
    \noindent \textbf{Comparison with existing methods:}
    We compared registration performance to VOTUS+Z\cite{VOTUS}, Harris-PIIFD+Z\cite{PIIFD}, GFEMR+Z\cite{GFEMR}, GDB-ICP+Z\cite{GDB}, 3D-SGR\cite{seg3D}, SRWCR\cite{SRWCR}. The first four are well-established  fundus registration methods for human eyes, 
    but here with parameters adapted for mouse images. The 3D-SGR is applied with mild adaption for the mouse dataset. The SRWCR is applied directly to the B-scans and then uses a warped volume to construct projections for evaluation. The overall results are listed in Table \ref{tab:regresult} and an example of registered OCT images is shown in Fig.\ref{fig:results}. Our proposed OT-GM combined with AWVGD and CD outperforms the other traditional methods with 72\% accuracy and 1.469 GC. Although other methods could prove valuable and perform reasonably well when applied to good quality images, they generally fail in mouse datasets. The average execution time for the proposed method for a pair of OCT volumes is 1131 seconds. The pre-processing step, x-y plane registration step and z-direction registration step take 398 seconds, 698 seconds and 35 seconds, respectively. Although not the fastest, the average time is reasonable and the accuracy is high. The code was implemented in MATLAB R2019 in Intel Core i7-8700 CP5 of 3.20GHz and 16 GB of RAM.
        \begin{table}[!t]
        \centering
        \caption{Successful Registration Accuracy, Average GC, and Run Time of Different Methods}
        \label{tabaccuracy}
        \begin{tabular}{|l|c|c|c|c|}
        \hline  
         \multicolumn{2}{|c|}{Method} & Accuracy(\%) & Aver GC & Time(s)\\
        \hline \hline        
        \multicolumn{2}{|c|}{SRWCR} & 0 & 0 & 1257 \\
        \hline        
        \multicolumn{2}{|c|}{3D-SGR} &  0 & 0 & 480 \\
        \hline         
        \multicolumn{2}{|c|}{GDB-ICP+Z} & 0 & 0 & 248  \\
        \hline        
        \multicolumn{2}{|c|}{GFEMR+Z} & 0 & 0 & 191  \\ 
        \hline        
        \multicolumn{2}{|c|}{Harris-PIIFD+Z} & 19 & 1.1067 & \textbf{218}  \\
        \hline        
        \multicolumn{2}{|c|}{VOTUS+Z} & 23 & 1.1235 & 865  \\
        \hline        
        \multirow{2}{*}{OURS} &CD &  38 & 1.3037 & 1120\\
        \cline{2-5}
         & \textbf{AWVGD+CD} & \textbf{72} & \textbf{1.469} & 1131\\
         \hline
        \end{tabular}
        \label{tab:regresult}
        \end{table}
    
    When AWVGD and CD are used simultaneously, the system reaches its best performance with 72\% accuracy. It can be seen from Fig.\ref{fig:results} (b) and (c) that results with AWVGD+CD aligns more vessel when compared to CD alone. Although only using AWVGD fails for all pairs, the registration performance with AWVGD+CD outperforms CD from the range 7\%-100\% with only 0.9\% computational time added. The 100\% improvement means that with only CD, the registration fails, and among all 31 successfully registered pairs, ten pairs had 100\% improvement with AWVGD. The average accuracy of the CD alone is 38\% while adding AWVGD improves this by 34\%. 
    
    We also analyze pairs mismatched using our method. This is mainly because of the difficulties with blood vessel extraction and the absence of HR apex. Poor vessel segmentation due to poor image quality directly affects the graph generation, which leads to wrong matches between key points. Inaccurate HR apex information affects the extraction of centroids and the calculation for z directional translations, which hampers the final registration.

\section{CONCLUSION AND FUTURE WORK}
This paper introduced an automated longitudinal OCT volume registration framework based on OT-GM with newly designed descriptors (AWVGD and CD). The results are evaluated both subjectively and objectively. Experimental results demonstrate the method outperforms other well-established state-of-the-art methods on mouse retinal images with uveitis with feasible execution time. In the future, we expect the framework to be extended to multi-modality (e.g.OCT-fundus, OCT-confocal) mouse retinal image registration.

\bibliographystyle{unsrt}  
\bibliography{3DOCTreg}
\end{document}